\documentclass[sigconf]{acmart}
\AtBeginDocument{%
  }

\copyrightyear{2026}
\acmYear{2026}
\setcopyright{cc}
\setcctype{by}
\acmConference[GECCO Companion '26]{Genetic and Evolutionary Computation Conference}{July 13--17, 2026}{San Jose, Costa Rica}
\acmBooktitle{Genetic and Evolutionary Computation Conference (GECCO Companion '26), July 13--17, 2026, San Jose, Costa Rica}
\acmDOI{10.1145/3795101.3805456}
\acmISBN{979-8-4007-2488-6/2026/07}

\begin{document}

\title{Motif Diversity in Human Liver ChIP-seq Data Using MAP-Elites}

\author{Alejandro Medina}
\email{alejandro_medina1@baylor.edu}
\orcid{0009-0005-7618-4947}
\affiliation{%
  \institution{Baylor University}
  \city{Waco}
  \state{Texas}
  \country{USA}
}

\author{Mary Lauren Benton}
\email{marylauren_benton@baylor.edu}
\orcid{0000-0002-5485-1041}
\affiliation{%
  \institution{Baylor University}
  \city{Waco}
  \state{Texas}
  \country{USA}
}

\renewcommand{\shortauthors}{Medina et al.}

\begin{abstract}

    Motif discovery is a core problem in computational biology, traditionally formulated as a likelihood optimization task that returns a single dominant motif from a DNA sequence dataset. However, regulatory sequence data allow multiple plausible motifs, reflecting underlying biological heterogeneity. In this work, we frame motif discovery as a quality-diversity problem and apply the MAP-Elites algorithm to evolve position weight matrix motifs under a likelihood-based fitness objective while explicitly preserving diversity across biologically meaningful dimensions.
    We evaluate MAP-Elites using three complementary behavioral characterizations that capture trade-offs between motif specificity, compositional structure, coverage, and robustness. Experiments on human CTCF liver ChIP-seq data aligned to the human reference genome compare MAP-Elites against a standard motif discovery tool, MEME, under matched evaluation criteria across stratified dataset subsets. Results show that MAP-Elites recovers multiple high-quality motif variants with fitness comparable to MEME’s strongest solutions while revealing structured diversity obscured by single-solution approaches. 

\end{abstract}




\maketitle

\section{Introduction} 
    
    DNA sequence motif discovery is a central problem in biology, where the goal is to identify short sequence patterns associated with functional roles such as transcription factor binding, splicing, or regulation \cite{stormo2000Binding}. Accurate motif identification is valuable for the interpretation of high-throughput sequencing assays, including chromatin immunoprecipitation sequencing (ChIP-seq), and is essential for understanding gene regulation \cite{park2009ChIP}.
    
    Most classical motif discovery methods formulate the problem as a likelihood-based optimization task, seeking a single motif model that best explains the observed data. Representative methods, such as Multiple EM for Motif Elicitation (MEME), which are highly effective but inherently designed to return only one or a small number of dominant motifs \cite{2010Gibbs, 2015MEMEsuite}. From a biological perspective, regulatory sequence data often exhibit heterogeneity: binding sites vary in specificity, degeneracy, and prevalence, and multiple motif variants may coexist within the same dataset \cite{stormo2000Binding,gupta2007SimilarityBetweenMotifs}. ChIP-seq studies have shown that combinatorial and context-dependent binding gives rise to multiple functional motif configurations rather than a single dominant pattern \cite{benner2010Macrophage}, motivating approaches that move beyond single optimal motifs.
    
    Evolutionary computation has been applied to DNA motif discovery, demonstrating that population-based stochastic search provides an alternative to probabilistic optimization \cite{2005MDGA,2009FitnessMotifs}. These approaches typically retain the objective structure of classical methods, converging toward a single best motif under a predefined fitness criterion \cite{2007populationClusteringMotif}. In this work, we frame motif discovery as a quality-diversity (QD) problem, aiming to discover diverse sets of high-quality motifs rather than a single optimal solution \cite{pugh2016QualityDiversity}.
    
    We use MAP-Elites to evolve position weight matrix (PWM) motifs under a likelihood-based fitness while preserving diversity across biologically meaningful behavioral characterizations \cite{mouret2015MAP-Elites}. Rather than returning a single motif, the resulting archive of high-quality motifs reveals trade-offs between motif specificity, prevalence, and structure. We evaluate three complementary descriptor pairings on sequences derived from human CCCTC-binding factor (CTCF) ChIP-seq data generated by the ENCODE Project \cite{2012ENCODE}, using stratified subsets to assess robustness across samples and comparisons to MEME as a widely adopted baseline \cite{2015MEMEsuite}. Rather than proposing a replacement for classical motif discovery tools, this work positions quality-diversity search as an exploratory complement, enabling structured exploration of motif landscapes that are otherwise collapsed into single consensus solutions. As a companion contribution, we emphasize illumination and interpretability over exhaustive validation. We further validate discovered motifs against curated transcription factor databases to confirm that the recovered diversity reflects biologically meaningful variation rather than optimization artifacts.  

%

\begin{figure*}[t]
    \centering
    \begin{minipage}[t]{0.33\textwidth}
        \centering
        \includegraphics[width=\linewidth]{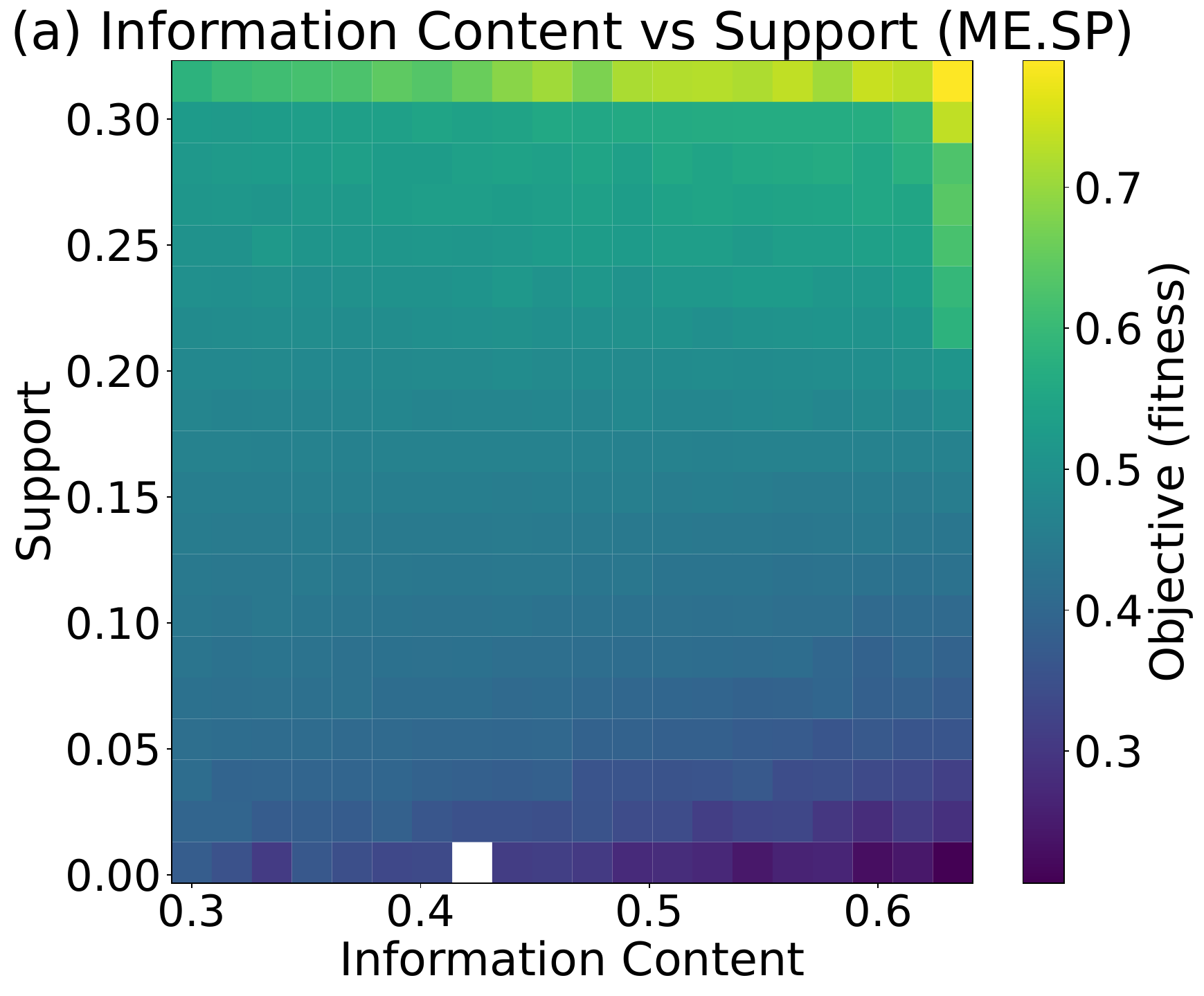}
    \end{minipage}
    \begin{minipage}[t]{0.33\textwidth}
        \centering
        \includegraphics[width=\linewidth]{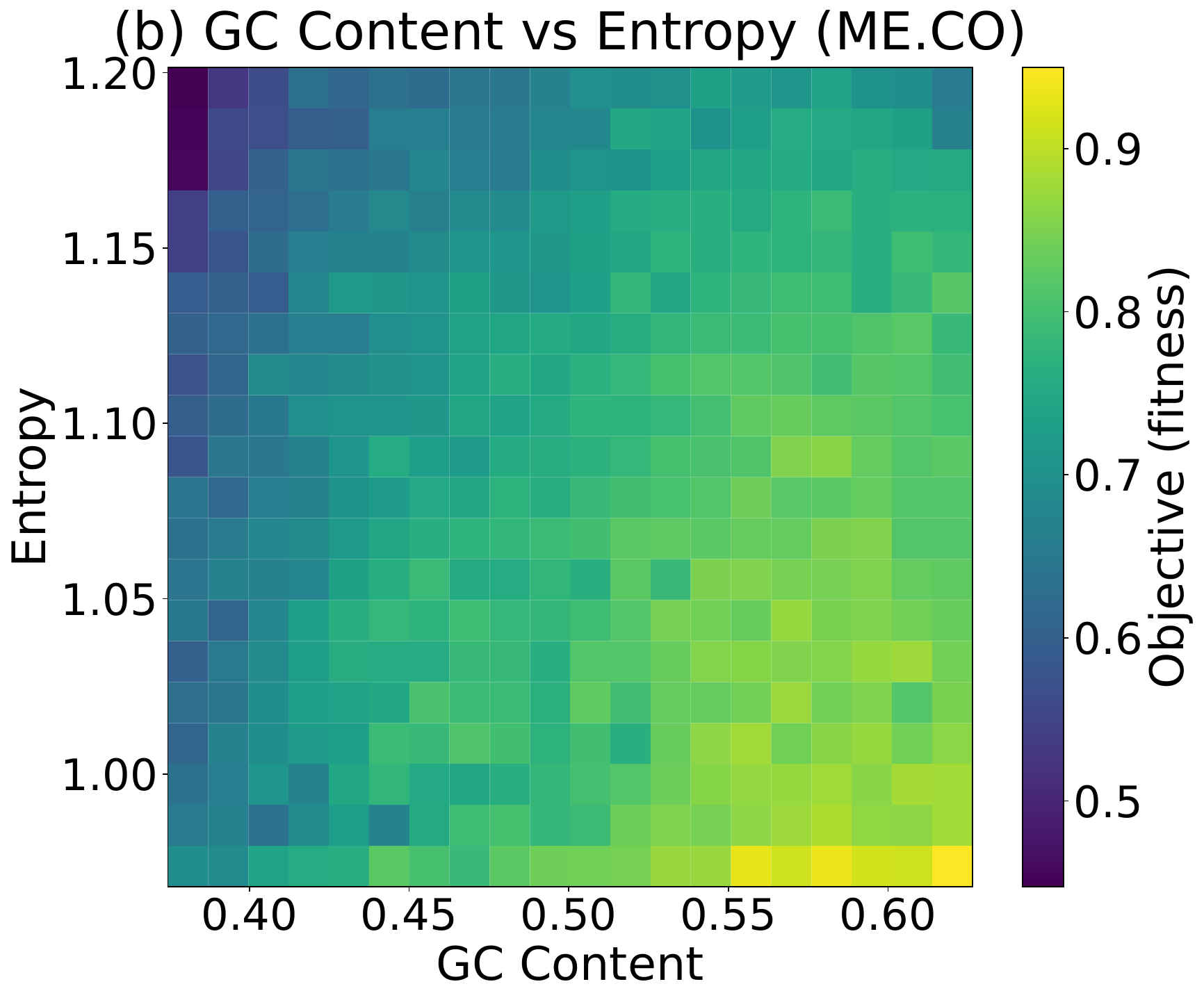}
    \end{minipage}
    \begin{minipage}[t]{0.33\textwidth}
        \centering
        \includegraphics[width=\linewidth]{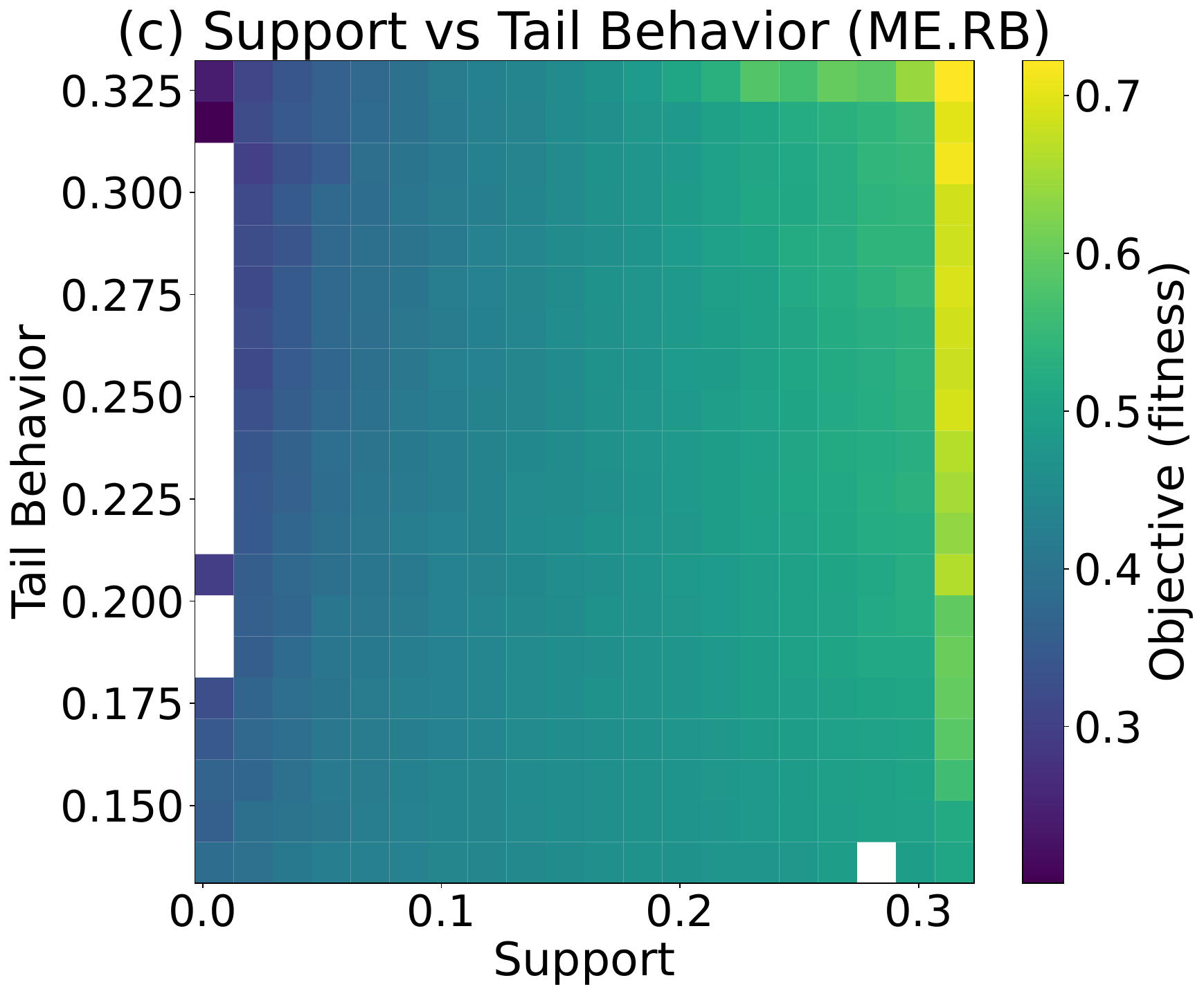}
    \end{minipage}

    \caption{MAP-Elites archive structure for a representative CTCF ChIP-seq subset. Heatmaps show elite motif fitness under three behavioral characterizations; white cells indicate unpopulated regions of the descriptor space.}
    \label{fig:archives}
\end{figure*}

\section{Methods}

    \subsection{Motif Representation}
        Candidate motifs are represented as fixed-length position weight matrices
        (PWMs) over the DNA alphabet $\{A,C,G,T\}$. Each motif is a row-stochastic
        matrix $P \in \mathbb{R}^{L \times 4}$, where rows define categorical
        nucleotide distributions at each position, consistent with standard motif
        discovery practice and comparable to MEME outputs
        \cite{stormo2000Binding,2015MEMEsuite}. Initial motifs are sampled from a
        symmetric Dirichlet distribution, with $L$ fixed across experiments. Variation operators preserve validity via logit perturbations with row-wise softmax or additive noise with clip-and-renormalize, ensuring row-stochastic PWMs.

    \subsection{Fitness Evaluation}

        A PWM is evaluated by its ability to discriminate foreground sequences from background using a log-odds scoring framework \cite{stormo2000Binding, stormo2013Modeling}. Let $M \in [0,1]^{L\times 4}$ denote a PWM of length $L$, $P_{\mathrm{bg}} \in [0,1]^4$ an empirical background nucleotide distribution, and $S$ a set of input sequences. For a length-$L$ window $w=(w_1,\dots,w_L)$, the log-odds score is
        \begin{equation}
            \ell(M,w) = \sum_{j=1}^L \log \frac{P_M(w_j \mid j)}{P_{\mathrm{bg}}(w_j)}.
        \end{equation}

    Given a sequence $s$, the best-hit score is computed as the maximum log-odds score over all valid windows and both strands, normalized by motif length: 
    \begin{equation}
        g(M,s) = \frac{1}{L} \max_{i} \max_{d \in \{+,-\}}
        \ell(M, s^{(d)}_{i:i+L-1}).
    \end{equation}

    The fitness of a motif $M$ is the average best-hit score over the top-$k$ sequences:
    \begin{equation}
    \label{eq:fitness}
        f(M) = \frac{1}{k} \sum_{s \in \mathrm{Top}_k(S; g)} g(M,s),
    \end{equation}  
    where $\mathrm{Top}_k(S; g)$ denotes the $k$ sequences in $S$ with the
    largest $g(M,s)$ values. In our experiments, $k = \left\lceil 0.2 \lvert S \rvert \right\rceil$, with a small trimmed variant (trim fraction $0.1$) for robustness.
    
    \subsection{MAP-Elites}
        We use MAP-Elites as the optimization framework for motif discovery. The algorithm maintains an archive indexed by behavioral descriptors, where each cell stores the highest-fitness motif discovered for that region of descriptor space \cite{mouret2015MAP-Elites}. Candidate motifs are generated through emitter-driven stochastic variation and compete locally within archive cells based on fitness, while diversity is preserved globally by construction. The resulting archive enables post hoc analysis of trade-offs between motif quality, specificity, prevalence, and structural variability, facilitating richer comparison with classical single-solution motif discovery methods.

\section{Experiments}
    All experiments are conducted on human CTCF ChIP-seq data aligned to the hg38 genome assembly \cite{2012ENCODE}. Candidate motifs are represented as fixed-length PWMs of length $L=19$, corresponding to the known length of the canonical CTCF binding motif \cite{jaspar2026, 2012ENCODE}. Unless otherwise stated, all algorithmic components other than the behavioral characterizations are held constant. 
    
    \subsection{Benchmark Dataset and Subset Construction} 
        Evaluating candidate motifs requires scanning each motif across all sequences, which is computationally expensive for larger datasets. To balance biological realism with computational tractability, the dataset is partitioned into five disjoint subsets, each containing an equal number of foreground sequences sampled from high-confidence ChIP-seq peaks. For each foreground set, a matched background set is constructed from genomic regions with similar characteristics. Both MAP-Elites and the baseline method MEME are evaluated independently on each disjoint subset one time. MEME is run on each subset to generate PWMs, which are then evaluated using the same likelihood-based fitness function (Equation~\ref{eq:fitness}) applied to MAP-Elites motifs, ensuring direct and fair comparison across methods. Subset-level evaluation enables assessment of robustness across samples drawn from the same underlying biological signal.

    \subsection{Behavior Characterizations}
        To isolate the effect of descriptor choice, all experiments use the same fitness function (Eq.~\ref{eq:fitness}), evolutionary operators, emitter configuration, archive resolution, and run length, while varying only the behavioral descriptors defining the MAP-Elites archive. We evaluate three complementary descriptor pairings, each designed to illuminate a distinct aspect of motif diversity: 
        
        \begin{enumerate}
            
            \item \textbf{Information Content and Support}

                Information Content captures motif specificity relative to background frequencies, while Support measures the fraction of foreground sequences in which the motif appears above a calibrated threshold \cite{stormo2000Binding}. This pairing distinguishes rare but highly specific motifs from broader patterns. 
        
            \item \textbf{GC Content and Entropy}
        
                GC Content measures average nucleotide composition, while Entropy captures the intrinsic positional variability independent of background frequencies \cite{stormo2000Binding}. This pairing emphasizes structural and compositional diversity among motifs.
        
            \item \textbf{Support and Score Tail Behavior}
        
                This pairing combines motif prevalence with a summary of the upper tail of best-hit score distributions, distinguishing motifs that moderately explain many sequences from those that strongly explain a smaller subset.
    
        \end{enumerate}
    
        Across all experiments, only behavioral characterizations are varied. Descriptor ranges are estimated automatically by evaluating randomly generated PWMs and setting bounds to the $[1\%, 99\%]$ quantiles over $400$ samples with a $10\%$ padding factor, avoiding manual tuning while ensuring biologically plausible archive discretization. For conciseness, the three descriptor pairings are denoted ME.SP (Information Content and Support), ME.CO (GC Content and Entropy), and ME.RB (Support and Tail Behavior).

    \subsection{Experimental Parameters}
        Fitness is computed using the likelihood-based objective in Equation~\ref{eq:fitness}. Because the score reflects relative foreground-background discrimination, fitness values may be negative, indicating weak or non-discriminative enrichment rather than invalid solutions. 
        
        Support thresholds are calibrated independently for each subset using background sequences, with thresholds set at the 95th percentile of background best-hit scores. MAP-Elites is implemented using the pyribs library with a GridArchive and an IsoLineEmitter \cite{2023Pyribs}. Each run executes for $1{,}000$ generations with a $20\times20$ archive, generating batches of $32$ candidates per iteration using isotropic and directional perturbations ($\sigma_{\text{iso}}=0.12$,  $\sigma_{\text{line}}=0.25$).
        
        Throughout each run, archive statistics including coverage, best fitness, and QD score are logged, and the full archive is retained for downstream analysis and comparison with MEME. In addition to peak fitness, these metrics characterize search dynamics and representational capacity. Archive coverage quantifies how much of the descriptor space is populated by viable motifs, while the QD score aggregates fitness across occupied cells, reflecting the ability of MAP-Elites to maintain simultaneously competitive solutions rather than converging to a single optimum.

\begin{table}[h]
            \centering
            \caption{Comparison of MEME and MAP-Elites: maximum fitness and mean $\pm$ standard deviation across subsets ($n=5$).}
            \label{tab:summary-results}
            \begin{tabular}{lcc}
            \toprule
            \textbf{Method} & \textbf{Max Fitness} & \textbf{Avg Fitness (mean $\pm$ std)} \\
            \midrule
            MEME     & $1.11$ & $-2.02 \pm 1.03$  \\
            ME.SP    & $0.883$ & $0.464 \pm 0.010$ \\
            ME.CO    & $0.950$ & $0.745 \pm 0.029$ \\
            ME.RB    & $0.797$ & $0.458 \pm 0.010$ \\
            
            \bottomrule
            \end{tabular}
        \end{table} 

\section{Results}
    Our evaluation examines how MAP-Elites recovers multiple high-quality motif hypotheses and how descriptor choice shapes archive structure and discovered solutions. Across all configurations and subsets, MAP-Elites achieves high archive coverage and stable QD scores, indicating effective maintenance of multiple competitive motifs rather than convergence to a single optimum.

\begin{figure}[h]
\centering


\begin{minipage}[t]{\columnwidth}
    \centering
    \includegraphics[width=\linewidth]{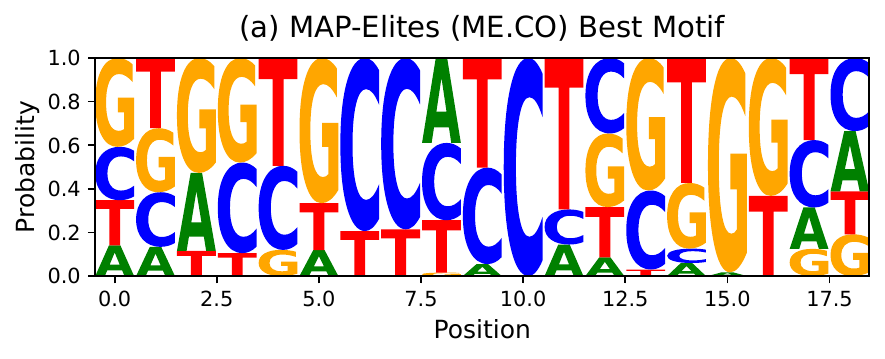}\\
\end{minipage}



\begin{minipage}[t]{\columnwidth}
    \centering
    \includegraphics[width=\linewidth]{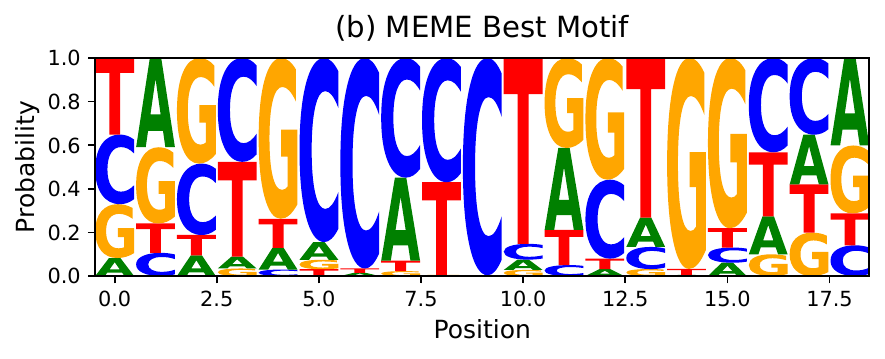}\\
\end{minipage}
\caption{Representative motif logos: highest-fitness ME.CO motif (a) and MEME motif (b) from Table~\ref{tab:summary-results}.
}
\Description{} 
\label{fig:motif-logos}
\end{figure}

    \subsection{Comparison with MEME}
        MEME identifies a single dominant motif per subset, with subsequent motifs degrading as residual signal is extracted. In contrast, MAP-Elites maintains a diverse set of high-quality motifs across descriptor-defined regions. As shown in Table~\ref{tab:summary-results}, MAP-Elites recovers multiple motifs with fitness comparable to MEME’s strongest solutions while exhibiting more stable average performance. The key distinction is representational capacity: MEME yields a single hypothesis, while MAP-Elites exposes structured families of competitive motifs.

        Figure~\ref{fig:motif-logos} shows the highest-fitness ME.CO motif alongside the MEME motif. These two are shown as the closest matches to canonical CTCF binding structure; additional MAP-Elites motifs from other descriptor spaces are omitted for space but included in quantitative and biological analysis.

        To assess biological validity, motifs were compared against the JASPAR CORE vertebrate database using Tomtom (MEME Suite motif comparison tool) \cite{gupta2007SimilarityBetweenMotifs,jaspar2026}. The MEME motif produced the strongest match to known CTCF profiles (best q-value $\approx 2.8 \times 10^{-12}$, overlap up to 19), while the best ME.CO motif also matched canonical CTCF with high significance (q-value $\approx 4.1 \times 10^{-10}$, overlap 15 to 16). Motifs from ME.SP and ME.RB likewise aligned to CTCF, though with weaker similarity (ME.SP: $10^{-7}$ to $10^{-5}$; ME.RB: $10^{-5}$ to $10^{-2}$). These results confirm that recovered motifs reflect CTCF binding structure rather than optimization artifacts.
    
    \subsection{Archive Structure and Descriptor Effects}
    We analyze how MAP-Elites organizes motif space under each descriptor by examining archive coverage, fitness distributions, and spatial structure. All configurations produce well-populated archives with smooth fitness gradients, indicating meaningful structure, but differ in how motif variation is organized. 
    
    \textbf{ME.SP.}
    In the ME.SP characterization, archive fitness is strongly dominated by support, with information content acting as a secondary refinement. As shown in Figure~\ref{fig:archives}(a), high-fitness motifs concentrate in regions of broad support, producing a stable average archive fitness of $0.464 \pm 0.010$ across subsets.
    
    \textbf{ME.CO.}
    ME.CO yields both the highest maximum fitness and the highest average fitness, albeit with higher variance across subsets (Table~\ref{tab:summary-results}). The corresponding archive (Figure~\ref{fig:archives}(b)) exhibits a pronounced fitness ridge aligned with GC-rich, low-entropy motifs, consistent with known compositional properties of CTCF binding sites. Although this behavioral characterization does not explicitly encode coverage, it recovers a compact and interpretable set of high-quality motifs that align closely with biological expectations.
    
    \textbf{ME.RB.}
    This characterization exposes a trade-off between coverage and robustness. As illustrated in Figure~\ref{fig:archives}(c), high-fitness motifs balance broad support with moderate score dispersion, whereas extreme tail behavior is consistently associated with lower-quality solutions.
    
    Taken together, these results show that while MAP-Elites achieves comparable coverage across descriptor choice, descriptor selection strongly shapes archive organization and motif variation. MAP-Elites not only recovers motifs consistent with known CTCF profiles, but organizes them into structured families reflecting biologically meaningful trade-offs rather than algorithmic artifacts.

\section{Conclusion}
    In this work, we formulate DNA motif discovery as a quality-diversity problem and show that illumination-based search recovers diverse, high-quality motif hypotheses from real biological sequence data. Using MAP-Elites with biologically meaningful behavioral characterizations, motif discovery can be structured around explicit trade-offs between coverage, compositional structure, and robustness rather than collapsing to a single optimum. On human CTCF ChIP-seq data, MAP-Elites recovers motifs with fitness comparable to MEME while maintaining a diverse set of alternative variants, capturing variability relevant to regulatory function. These results further demonstrate that descriptor choice shapes not only aggregate performance but also the organization and interpretability of the motif landscape, which is inaccessible to classical single-solution approaches.  
    
    This study is an initial step and has certain limitations. We focus on a single dataset, fixed motif length, and a single illumination algorithm to isolate descriptor and archive effects; comparisons to alternative evolutionary or multi-objective methods, variable-length motif discovery, and broader biological validation against reference motif databases are left for future work. While some descriptors may be statistically related, the high archive coverage and structured fitness landscapes observed here suggest sufficient diversity for effective illumination; more formal analysis of descriptor interactions remains an open direction. In addition, the fitness function is tailored to foreground-background discrimination. Extending this framework to additional datasets, alternative QD algorithms, descriptor analyses, and fitness formulations is a natural next step. Overall, these results position quality diversity as a promising framework for exploratory motif analysis and motivate future work that scales illumination-based approaches across transcription factors, datasets, and biological contexts. 
    
\begin{acks}
    Generative AI was used to assist with LaTeX figure formatting.
\end{acks}

\bibliographystyle{ACM-Reference-Format}
\bibliography{references}

\end{document}